\documentclass[10pt,twocolumn,letterpaper]{article}

\let\svthefootnote\thefootnote
\newcommand\blankfootnote[1]{%
  \let\thefootnote\relax\footnotetext{#1}%
  \let\thefootnote\svthefootnote%
}
\let\svfootnote\footnote
\renewcommand\footnote[2][?]{%
  \if\relax#1\relax%
    \blankfootnote{#2}%
  \else%
    \if?#1\svfootnote{#2}\else\svfootnote[#1]{#2}\fi%
  \fi
}

\usepackage[pagenumbers]{cvpr} 

\usepackage{amsmath,amsfonts,bm}

\def\eqref#1{equation~\ref{#1}}

\def\1{\bm{1}}

\DeclareMathAlphabet{\mathsfit}{\encodingdefault}{\sfdefault}{m}{sl}
\SetMathAlphabet{\mathsfit}{bold}{\encodingdefault}{\sfdefault}{bx}{n}

\usepackage[inline]{enumitem}
\usepackage{multirow}
\usepackage{subcaption}
\usepackage{xcolor}
\usepackage[subtle]{savetrees}
\usepackage{graphicx}
\usepackage{booktabs}
\usepackage[accsupp]{axessibility}  

\usepackage[pagebackref,breaklinks,colorlinks]{hyperref}
\let\footnote\thanks

\usepackage[capitalize]{cleveref}
\crefname{section}{Sec.}{Secs.}
\Crefname{section}{Section}{Sections}
\Crefname{table}{Table}{Tables}
\crefname{table}{Tab.}{Tabs.}

\def\cvprPaperID{67} 
\def\confName{EarthVision}
\def\confYear{2022}

\begin{document}

\title{Fast building segmentation from satellite imagery and few local labels}

\author{Caleb Robinson\\
Microsoft AI for Good Research Lab\\
{\tt\small caleb.robinson@microsoft.com}
\and
Anthony Ortiz\\
Microsoft AI for Good Research Lab\\
{\tt\small anthony.ortiz@microsoft.com}
\and
Hogeun Park\\
World Bank\\
{\tt\small hpark2@worldbank.org}
\and
Nancy Lozano Gracia\\
World Bank\\
{\tt\small nlozano@worldbank.org}
\and
Jon Kher Kaw\\
World Bank\\
{\tt\small jkaw@worldbank.org}
\and
Tina Sederholm\\
Microsoft AI for Good Research Lab\\
{\tt\small tinase@microsoft.com}
\and
Rahul Dodhia\\
Microsoft AI for Good Research Lab\\
{\tt\small radodhia@microsoft.com}
\and
Juan M. Lavista Ferres\\
Microsoft AI for Good Research Lab\\
{\tt\small jlavista@microsoft.com}
}
\maketitle

\begin{abstract}
Innovations in computer vision algorithms for satellite image analysis can enable us to explore global challenges such as urbanization and land use change at the planetary level. However, domain shift problems are a common occurrence when trying to replicate models that drive these analyses to new areas, particularly in the developing world. If a model is trained with imagery and labels from one location, then it usually will not generalize well to new locations where the content of the imagery and data distributions are different. In this work, we consider the setting in which we have a single large satellite imagery scene over which we want to solve an applied problem -- building footprint segmentation. Here, we do not necessarily need to worry about creating a model that generalizes past the borders of our scene but can instead train a local model. We show that surprisingly few labels are needed to solve the building segmentation problem with very high-resolution (0.5m/px) satellite imagery with this setting in mind. Our best model trained with just 527 sparse polygon annotations (an equivalent of $1500 \times 1500$ densely labeled pixels) has a recall of 0.87 over held out footprints and a R2 of 0.93 on the task of counting the number of buildings in $200 \times 200$ meter windows. We apply our models over high-resolution imagery in Amman, Jordan in a case study on urban change detection.
\end{abstract}

\begin{figure}[t]
    \centering
    \includegraphics[width=0.9\linewidth]{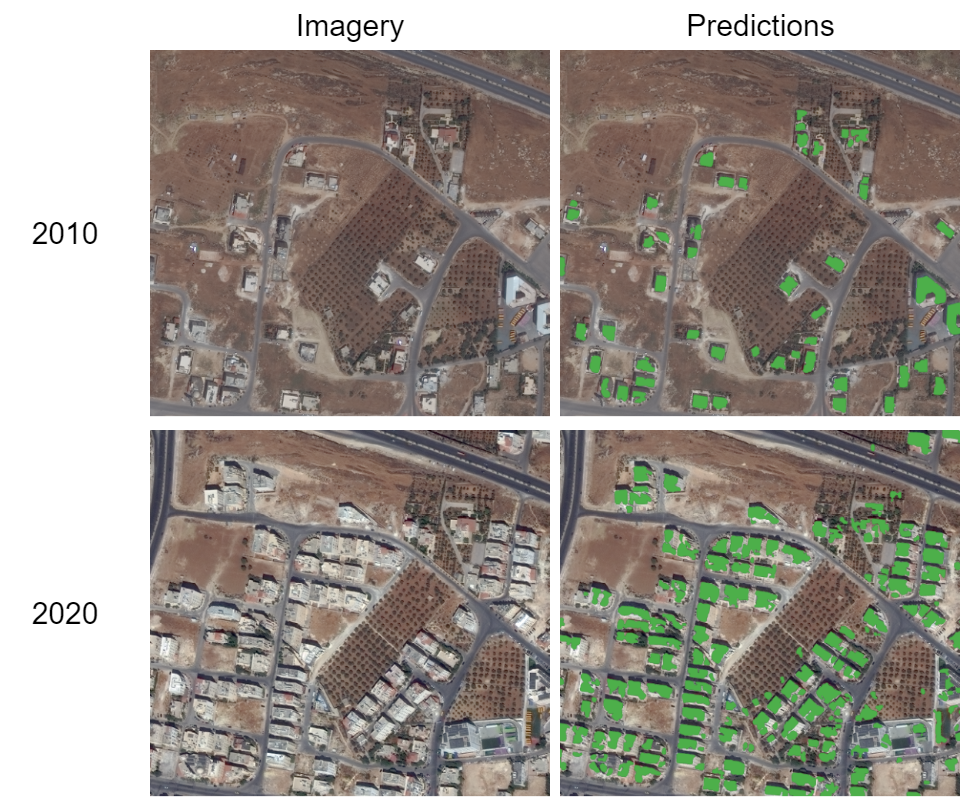}
    \caption{Example predictions from building footprint segmentation models trained from scratch using only high-resolution RGB satellite imagery and $< 600$ polygon labels. The imagery shows a rapidly developing part of Amman, Jordan in 2010 and 2020. Predictions are made with models trained using 527 and 367 sparse polygon labels collected from a small part of the 2010 and 2020 scenes respectively.}
    \label{fig:teaser}
\end{figure}

\blankfootnote{The authors declare that the findings, interpretations, and conclusions expressed in this paper are entirely those of the authors. They do not necessarily represent the views of their affiliated organizations.}

\section{Introduction}
\label{sec:intro}

Building footprints are a crucial piece of data in many applications, for example: disaster response damage assessments, population mapping, and urban planning \cite{gupta2019xbd,xu2018building}. High-resolution satellite or aerial imagery can be hand-labeled to create accurate building footprints, however such an approach quickly becomes prohibitively expensive as the scale of the problem increases. Model-based approaches that predict building footprints from satellite imagery can easily be applied at large scales, however must deal with problems of generalizing across different imaging conditions and geographies to produce meaningful results.

Two approaches are traditionally used to improve the generalization performance of such models: domain adaptation methods, and training with large diverse training sets.
First, domain adaptation (DA) methods have been proposed for Earth Observation applications~\cite{tuia2021recent} to account for \textit{distribution shifts} in satellite imagery and create generalizable models. In general, DA methods aim to obtain models robust to differences between two different datasets (a source and target) with different image and label distributions. DA methods tackle the domain shift issues by either aligning the distribution of the source set to match that of the target set~\cite{ammour2018asymmetric,murez2018image,deng2021scale} or by mapping both sets to a common space. Deng et al. proposed a scale aware adversarial learning framework for domain adaptation in building segmentation between datasets collected at different scales~\cite{deng2021scale}. Tasar et al. proposed to use a Generative Adversarial Network (GAN) to generate fake synthetic training images that are semantically similar to the training images, but with spectral distribution similar to the distribution
of the target imagery for better generalization of building segmentation models for different cities~\cite{tasar2020colormapgan}. These approaches tend to improve overall model performance in the target domain, however the results are often suboptimal compared to supervised learning in the target domain.
Second, training with larger diverse datasets can produce models with better generalization performance. Some existing open datasets for building damage assessment or building footprint segmentation, such as the xBD~\cite{gupta2019xbd} and SpaceNet datasets~\cite{van2018spacenet}, purposefully include labels from different geographic regions. This allows practitioners to experiment with domain adaptation methods and measure generalization performance. However, these datasets alone are not sufficient for training models that can be applied \textit{as-is} to new satellite imagery from around the Earth~\cite{benson2020assessing}.

Previous projects have shown positive results from building footprint segmentation models at continent level scales using a mixture of domain adaptation methods and large training datasets. For example, Google released a dataset of modeled building footprint estimates for Africa~\cite{sirko2021continental}, Microsoft has released building footprint datasets for the continental US, Canada, Australia, and South America~\cite{msftBuildingFootprints}, while Facebook has released population density datasets driven by building detection models for over 160 countries~\cite{fbHRSL}. The Google Open Buildings model required ``manually labelling 1.75 million buildings in 100k images''~\cite{sirko2021continental}, the Microsoft US Building footprint model was generated with access to ``millions of labels'' and used unsupervised learning techniques to improve generalization performance~\cite{msftUSBuildingFootprints}, while Facebook's building mapping efforts started with ``a seed dataset of around 1M labeled patches of imagery'' and used a variety of techniques to improve generalization performance~\cite{bonafilia2019building}.
While these projects have all resulted in useful \textit{datasets} that help advance humanitarian goals, the models and imagery behind these efforts have not been open-sourced and therefore cannot be used on new imagery. This is a limitation for applications such as \textit{building damage assessment after disasters} that require on-demand processing of new imagery and \textit{urban change detection} that require processing multiple layers of historical imagery on-demand (versus generating a static map of current buildings).

In this work, we come at the applied problem of building segmentation from a different direction and do not attempt to train models that can generalize to new inputs at all. Instead, we ask the question, ``how many labels do we need to collect from a single scene in order to train a model to an acceptable performance within that scene?''. We build a workflow in which we can quickly solicit polygon-based labels with a web application that are tied to a given satellite imagery scene, train a model using those labels and imagery, then run the model over the whole scene producing the desired predictions. Given new imagery, this workflow can be executed in less than a day of work to produce consistent results without concerns about out-of-domain generalization performance. We use this workflow in Amman, Jordan to create a detailed change map of buildings from 2010 to 2020 that can directly inform urban planning decisions. 

To summarize, our contributions include:
\begin{itemize}
    \item An analysis of the necessary components in an on-demand building segmentation modeling method with few sparse labels.
    \item An open-source web-based tool for collecting polygon labels over a given remotely sensed imagery scene: \url{https://github.com/microsoft/satellite-imagery-labeling-tool}.
    \item A case study applying our methods to urban change detection in Amman, Jordan.
\end{itemize}

\begin{figure*}[t]
    \centering
    \includegraphics[width=0.7\linewidth]{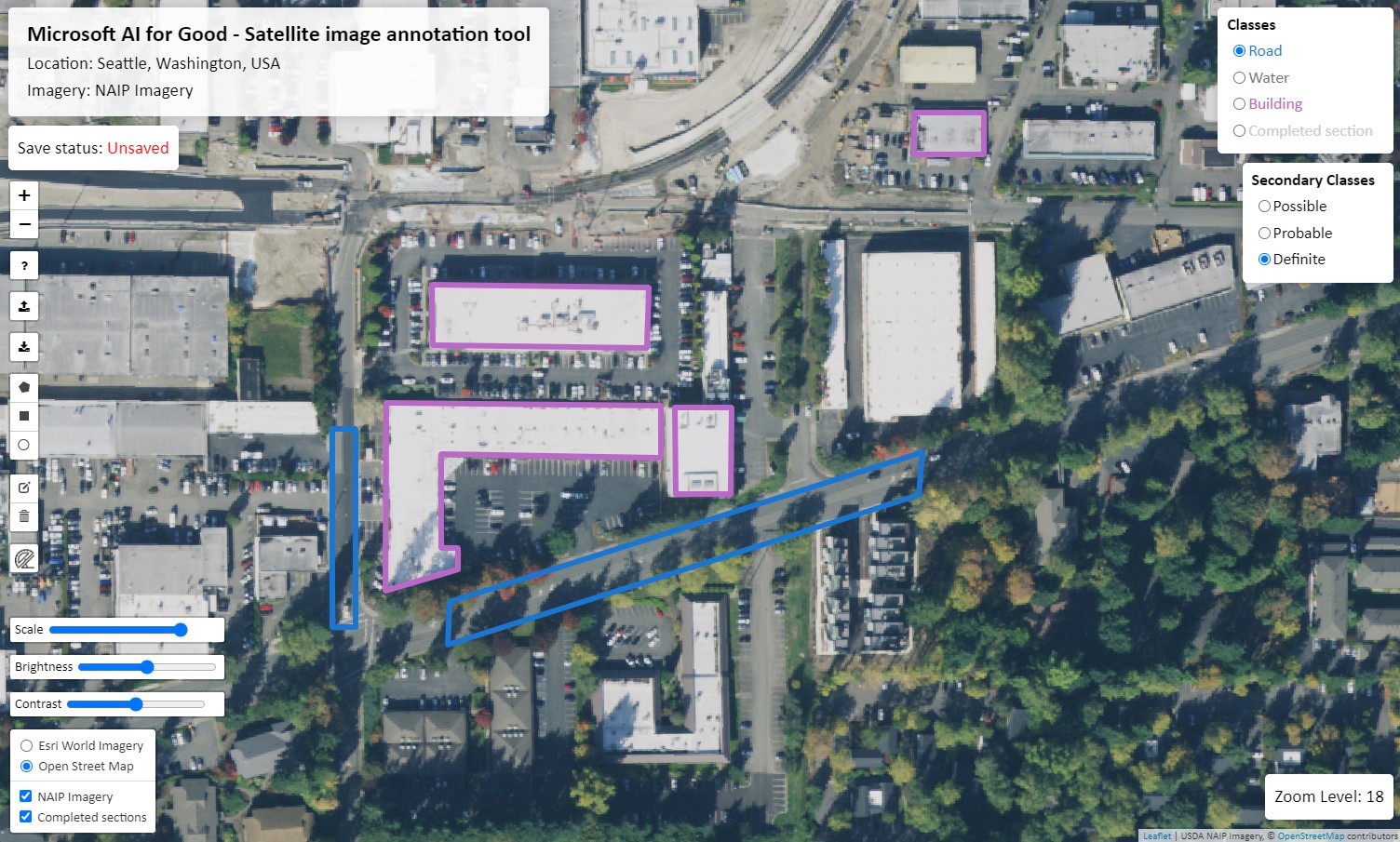}
    \caption{A web-based satellite imagery labeling tool allows us to quickly solicit polygon labels that are aligned with a particular satellite imagery scene. Users receive a link for a particular scene, then can annotate the imagery with different labels and confidence levels, and finally download the annotations in a GeoJSON format for further processing or distribution.}
    \label{fig:labeling-tool}
\end{figure*}

\begin{figure*}[th]
    \centering
    \includegraphics[width=0.75\textwidth]{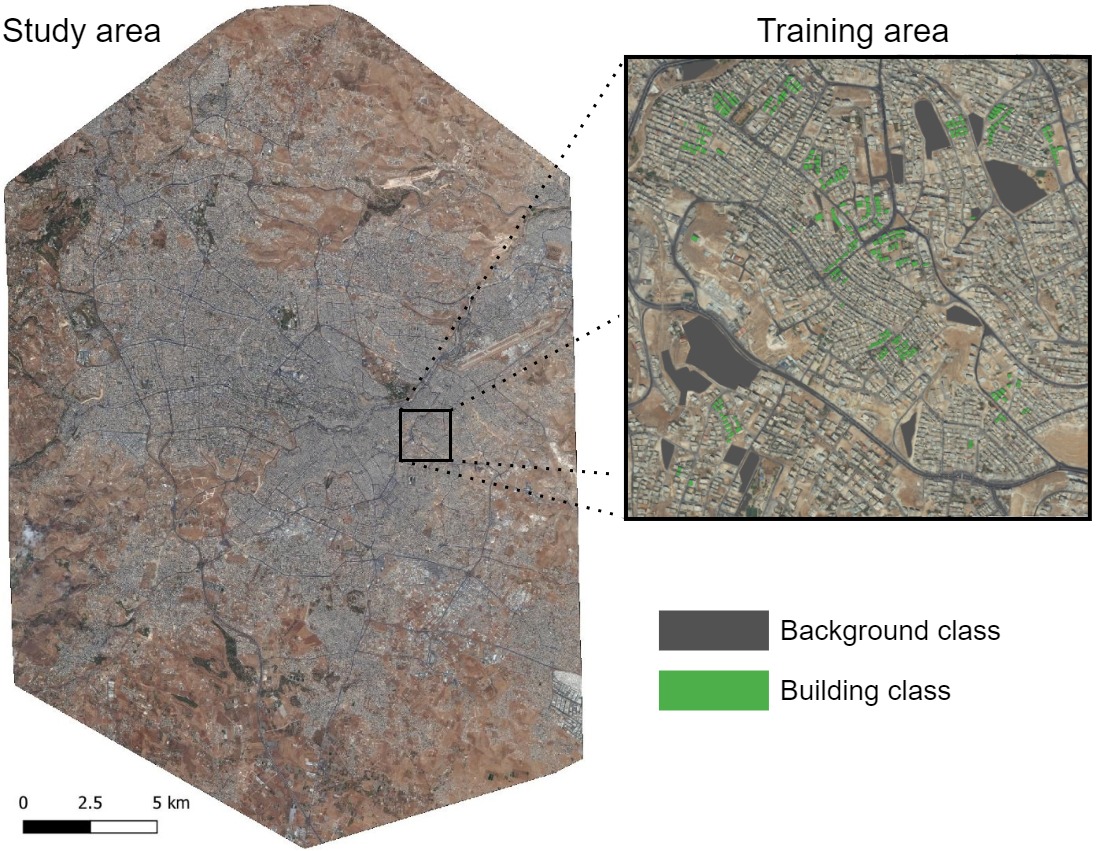}
    \caption{Overview of the Amman, Jordan study area showing the 2020 imagery layer. The map inset shows the area over which the \textbf{train} split labels were created along with the 367 training polygons for 2020.}
    \label{fig:overview-map}
\end{figure*}

\section{Problem statement}
Given a large satellite image, $\mathbf{X}$, and an accompanying sparse label mask, $\mathbf{Y}$, we want to learn the parameters of a semantic segmentation model, $f(\mathbf{X}; \theta) = \hat{Y}$ that can make dense predictions, $\hat{Y}$, over the entire image. Here, entries in the label mask $Y_{ij}$ indicate that the corresponding pixel in the imagery $X_{ij}$ is either part of a building footprint, is not part of a building footprint, or is unknown. We note that with modern commercial high-resolution satellite imagery, the spatial dimensions of $\mathbf{X}$ can exceed $100,000 \times 100,000$ pixels and cover hundreds of square kilometers. We assume that $\mathbf{X}$ was captured at a single point in time, i.e. that the only source of variance in the imagery that $f$ needs to capture is related to how buildings look in that particular scene (and not due other differences such as lighting, off nadir angles, seasonality, etc.). While in this work we specifically focus on the building segmentation problem, the same ideas can be applied to other earth observation tasks.

\section{Methods}

Our proposed methods are simple, to generate building footprints in a large high-resolution satellite imagery scene we 1.) sparsely label instances of buildings and non-building (or background) classes in that scene; 2.) train a semantic segmentation model from scratch using these labels; and 3.) run the trained model over the entire scene.

\subsection{Satellite imagery labeling tool} \label{subsec:annotation-tool}

Creating labels over large satellite imagery scenes is a non-trivial task. First, multispectral high-resolution satellite imagery scenes can be many gigabytes in size and therefore difficult to download locally without a dedicated internet connection. Second, these scenes usually require domain expertise and specialized desktop GIS software to visualize and annotate (for example, a user may need to select the RGB bands in the imagery and perform normalization steps to convert the spectral values into a reasonable range for visualization). Finally, once a user has created annotations they must be exported and distributed with their geographic metadata to avoid alignment issues between the annotations and imagery.

Considering these difficulties, we have created an open-source web-based tool that allows users to create annotations over basemap imagery layers, then export the annotations in a GeoJSON format for use in modeling pipelines (Figure \ref{fig:labeling-tool} shows a screenshot of the tool). This allows a technical user with GIS expertise to create a hosted basemap from a satellite imagery scene, then simply distribute a URL in order to solicit labels from non-technical annotators. We note that this same type of setup -- web-based annotation of satellite imagery -- is used in other hosted applications such as Open Street Map's online map editor. Our implementation, however, does not require any server components other than a standard web server (optionally, a tile server for dynamically rendering of basemaps). In other words, the entire application runs in a client's web browser. This simplifies the deployment of new instances of the application.

\subsection{Modeling} \label{subsec:modeling}

We treat the problem of creating building footprints as a standard semantic segmentation problem with an additional building polygonization step (i.e. converting pixel-wise predictions of buildings into polygons) and use a U-Net model architecture~\cite{ronneberger2015u} as implemented in the Segmentation models PyTorch library~\cite{yakubovskiy2019}.

We train models using an unweighted pixel-wise cross entropy loss that ignores pixels that haven't been labeled. Related work has proposed using weighting schemes that heavily weight the loss of pixels at the edges of building masks~\cite{ronneberger2015u,sirko2021continental,jiwani2021semantic}, using different loss functions such Focal Loss, Tversky loss, Focal-Tversky loss~\cite{sirko2021continental}, Jaccard loss~\cite{iglovikov2018ternausnetv2}, or F-Beta loss~\cite{jiwani2021semantic}, as well as deriving and predicting a ``building-edge'' class in a multi-task setting~\cite{iglovikov2018ternausnetv2}. These innovations focus on reducing false positive predictions, reducing the impact of imbalanced training samples, and, importantly, encouraging the network to predict background class labels between adjacent buildings (versus ``merging'' adjacent buildings). In contrast to these approaches, we found it sufficient to buffer our sparse building footprint labels with a ring of ``background'' labels. Specifically, we assume that every pixel within 2 meters of a ``building'' label (and not labeled otherwise) is a ``background'' label. We find that models trained with these buffered labels are able to separate buildings while models trained on the unbuffered sparse labels do not (see Section \ref{sec:results}). Finally, we train using an AdamW optimizer, we decay learning rate by a factor of 0.1 on training loss plateaus and we use random rotation, flipping, sharpness and color jitter augmentations as implemented by the Kornia library~\cite{riba2020kornia}. We tune learning rate, weight decay, and weight initialization hyper-parameters based on validation set performance.

In a post-processing ``polygonization'' step, we convert the per-pixel predictions made by our model into building polygons for use in downstream applications. Here, we assume we have used a trained model to run inference over a whole area classifying each pixel as ``building'' or not. First, we run a median filter with a $7 \times 7$ kernel to remove small erroneous predictions and fill gaps in predicted buildings. Next, we convert each connected component of building pixels into a polygon (where pixel borders are replaced by edges), and simplify the polygon with a tolerance of 0.5m using the Douglas-Peucker algorithm~\cite{douglas1973algorithms} as implemented by the shapely library~\cite{shapely}. Finally, we remove a polygon (replacing the predictions with ``background'') if it has an area that is less than 30 square meters.

\begin{table}[tb]
\centering
\resizebox{\linewidth}{!}{%
\begin{tabular}{@{}ccccc@{}}
\toprule
\textbf{Date} & \textbf{\begin{tabular}[c]{@{}c@{}}Off Nadir\\ Angle\end{tabular}} & \textbf{\begin{tabular}[c]{@{}c@{}}Sun\\ Elevation\end{tabular}} & \textbf{\begin{tabular}[c]{@{}c@{}}Max Target\\ Azimuth\end{tabular}} & \textbf{Area} \\ \midrule
6/4/2010 & 18.6° & 74.5° & 172.7° & \multirow{3}{*}{581.7 km²} \\
6/4/2010 & 11.5° & 74.6° & 49.8° &  \\
6/15/2010 & 13.4° & 74.7° & 61.1° &  \\
9/24/2020 & 27.8° & 52.2° & 298.9° & 558.6 km² \\ \bottomrule
\end{tabular}%
}
\caption{World View 2 imagery metadata from the scenes used over Amman.}
\label{tab:dataset-imagery}
\end{table}

\subsection{Datasets}

We use 4 separate high-resolution (0.5m/px) RGB satellite imagery scenes from Maxar's WorldView-2 satellite in this study, detailed in Table \ref{tab:dataset-imagery}. We create a mosaic covering Amman with the three scenes from June 2010, and use this as the \textbf{2010} layer, and use the single scene from September 2020 as the \textbf{2020} layer.

We use the labeling tool described in Section \ref{subsec:annotation-tool} to collect four sets of labels, or \textit{splits}, over each layer of imagery:
\begin{description}
    \item[Train] consists of sparse polygon labels created over the same $4\text{km}^2$ area from both layers. These labels include ``background'', ``road'', and ``building'' classes.
    \item[Val] consists of dense ``building'' class labels created over a subset of the $4\text{km}^2$ training area (that does not contain labels in the \textbf{train} split). Here, we label each building in a small area so that we can assume that all \textit{unlabeled} pixels in that area are not buildings (i.e. background or road). This allows us to measure the pixel-wise precision and recall for the ``building'' class during training.
    \item[Test] consists of sparse polygon ``building'' class labels created by randomly labeling buildings over the entirety of each layer. We hold these labels out during training and use them to approximately measure the recall of our models over the entire layers. Note that we cannot measure precision using this set of labels as knowing whether a prediction is a false positive requires knowing where \textit{all} true positives are. 
    \item[Test counts] consists of counts of the number of buildings over 29 randomly sampled $200 \times 200$ meter polygons from each layer. Similar to the \textbf{test} set, we hold these labels out during training. This allows us to measure how well our models capture the density of actual buildings in each layer. For example, a trivial model that labels every pixel as the ``building'' class would have 100\% recall, however the \textit{count} of the number of buildings predicted by this model would be completely uncorrelated with the actual number of buildings in an area.
\end{description}

Figure \ref{fig:overview-map} shows an overview of the study area and the small area used for training in Amman, Jordan for 2020. Notice how the training area is less than 1\% of the total study area with only 367 training polygons available. 

Table \ref{tab:dataset-composition} shows the number of labels of each class we collected over each layer. We differentiate between ``background'', ``road'', and ``building'' classes in order to test whether a finer grained classification can improve the performance of models on the building class. Finally, we estimate that an experienced user can create $\sim 6$ labels per minute (our average rate). At this rate it would take a total of $\sim 5$ hours of labeling effort to reproduce the \textbf{train}, \textbf{val} and \textbf{test} splits we use here. Reproducing the \textbf{test counts} dataset is similarly easy as it only involves counting and recording the number of buildings over a few fixed areas.

\begin{table}[tb]
\centering
\resizebox{\linewidth}{!}{%
\begin{tabular}{@{}cccccc@{}}
\toprule
\multicolumn{1}{c}{\textbf{Split}} & \multicolumn{1}{c}{\textbf{Layer}} & \multicolumn{1}{c}{\textbf{Background}} & \multicolumn{1}{c}{\textbf{Road}} & \multicolumn{1}{c}{\textbf{Building}} & \multicolumn{1}{c}{\textbf{Totals}} \\ \midrule
\multirow{2}{*}{\textbf{Train}} & 2010 & 59 & 31 & 437 & 527 \\
 & 2020 & 27 & 27 & 313 & 367 \\ \midrule
\multirow{2}{*}{\textbf{Val}} & 2010 & - & - & 40 & - \\
 & 2020 & - & - & 41 & - \\ \midrule
\multirow{2}{*}{\textbf{Test}} & 2010 & - & - & 344 & - \\
 & 2020 & - & - & 467 & - \\ \midrule
\multirow{2}{*}{\textbf{Test counts}} & 2010 & - & - & - & 29 counts \\
 & 2020 & - & - & - & 29 counts \\ \bottomrule
\end{tabular}%
}
\caption{Size of the different splits (in number of polygons) for the 2010 and 2020 layers over Amman.}
\label{tab:dataset-composition}
\end{table}

\subsection{Baseline methods and performance metrics}

We compare our modeling approach against off-the-shelf building segmentation models from~\cite{jiwani2021semantic} and to a random forest model from the scikit-learn library~\cite{scikit-learn}.

Jiwani et al. provide pre-trained weights for three models trained on the SpaceNet, CrowdAI, and Urban3D datasets~\cite{jiwani2021semantic}. Identical to our problem setting, these models are trained to segment buildings from RGB satellite imagery, and notably, were the \textit{only} publicly available models we could find for this task. We use these models \textit{as-is} to test how the best off-the-shelf approaches will work when applied on new imagery. That is, to simulate how a practitioner may try to use the models in an applied setting. We expect that these models would perform better if fine-tuned with labeled data from our two layers of imagery, and we expect that the methods proposed in~\cite{jiwani2021semantic} are relevant to our task however have left this investigation to future work.

The local random forest model uses the RGB values at a pixel as a feature representation and is trained over the labeled pixels in each layer's \textbf{train} split. We merge the road and background classes into a single class to reduce the problem to a binary classification. This serves as very simple color-based segmentation of the imagery and a lower bound on what should be possible for other approaches.

We test the predictions made by each method in three ways:
\begin{itemize}
    \item Pixel-wise F1 score of the building class using the dense labels from the \textbf{val} splits.
    \item Recall@$k$ using the building level labels from the \textbf{test} splits. We define Recall@$k$ as the percentage of buildings in which $\geq k$\% of the pixels in a building are correctly predicted as the building class. In other words, a labeled building in the \textbf{test} split is counted as a true positive if $k$\% of the pixels in that building are predicted as the building class, else it is counted as a false negative.
    \item Coefficient of determination (R2) calculated between the counts of buildings over the 29 areas in the \textbf{test count} splits compared to the predicted number of buildings over the same areas. Specifically, for each of the 29 $200 \times 200$ meter areas in the \textbf{test count} split, we compute the number of model predicted buildings by totaling the number of polygons that intersect with that area after the polygonization post-processing step described in Section \ref{subsec:modeling}. 
\end{itemize}

\begin{figure}[th]
    \centering
    \includegraphics[width=1\linewidth]{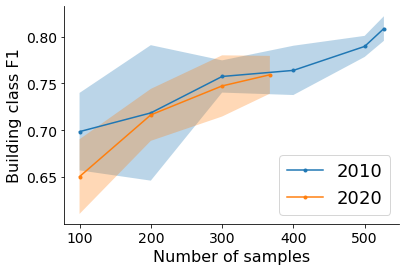}
    \includegraphics[width=1\linewidth]{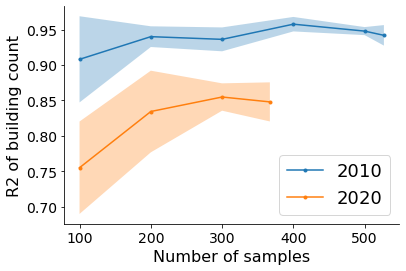}
    \caption{The effect of the number of training samples on U-Net model performance in the 2010 and 2020 layers. (\textbf{Top panel}) shows the impact on the building class F1 score as measured on the \textbf{val} splits. (\textbf{Bottom panel}) shows the impact on the R2 of the predicted building footprint counts as measured on the \textbf{test counts} splits.}
    \label{fig:samples}
\end{figure}

\section{Experiments and results} \label{sec:results}

Table \ref{tab:results-main} shows the performance of our best performing U-Net model trained with local labels along with the baseline method performance. We observe that the CrowdAI pretrained model and U-Net models are the only models that make reasonable predictions. Here, the CrowdAI model makes a large number of false positive predictions and frequently ignores boundaries between adjacent buildings in both layers of imagery. This results in high recall scores, but predictions that don't correlate as well with the actual number of buildings over an area -- achieving 0.76 and 0.67 R2 scores for 2010 and 2020 respectively. The U-Net, in comparison, has a slightly lower recall scores (86.9 vs. 90.7 and 83.7 vs. 89.1) but is able to consistently segment individual buildings. This results in predictions that \textit{are highly correlated with the actual number of buildings} -- achieving 0.93 and 0.84 R2 scores. The SpaceNet model achieves near perfect recall, but with a 0 R2 score, as it predicts the building class for almost any input. As expected, the random forest model does not perform well as per-pixel color features alone are not sufficient to identify buildings in high-resolution imagery.

We also investigate the impact of the number of labels used on the performance of the best U-Net model configurations. Here, we subsample different numbers of polygons without replacement and train the U-Net with the reduced set of labels. Figure \ref{fig:samples} shows the averaged performance over 5 different subsamples / random weight initializations (+/- 1 standard deviation is shown in the shaded area). We observe a consistent increase in building class F1 scores across both layers of imagery up to the maximum number of labels we gathered which suggests that more labels would continue to increase performance of the model locally and produce better footprints. However, we also observe an immediate plateau in the R2 counting performance of the models, and large standard deviations in performance. This suggests that even 100 to 200 labels is sufficient to train a building footprint model especially if the predictions made by that model are used primarily for building density estimation tasks. The fact that some sets of 100 labels achieve very high R2 values in the 2010 layer further suggest that \textit{which} labels are used in training are more important than how many are used. Overall, we find that the performance of the models on the 2020 layer is uniformly worse than on the 2010 layer. One explanation for this is that the 2020 imagery was taken with a greater off-nadir angle than the 2010 imagery (Table \ref{tab:dataset-imagery}) and is of generally lower quality.

Finally, ablation studies to investigate the effect of different modeling choices. Specifically, we test the impact of merging the ``road'' class with the ``background'' class or not, the impact of using ImageNet pre-trained weights versus random weight initialization, and the effect of \textit{buffering} the ``building'' class labels with a small ring of ``background'' class labels. We run each combination of these choices for several learning rate and weight decay parameters. Table \ref{tab:results-ablation} show the building class F1 scores for each choice averaged over all valid combinations (e.g. the score of 0.710 for ``Road class separate'' in 2010 is averaged over all other runs in the grid that were trained using ``road'' as a unique class). First, we observe that merging the road class with the background class results in slightly better performance in both layers of imagery. Second, we find that buffering the ``building'' class does not impact the results in 2010, however dramatically improves the performance in the more off-nadir 2020 imagery. Finally, we find that starting from a random weight initialization is \textit{better} than starting from ImageNet pre-trained weights, with a large improvement in 2010. We observe that the models that were trained starting from ImageNet weights are often able to overfit completely, achieving close to 0 training loss but producing nonsensical predictions elsewhere in the imagery.

\begin{table}[th]
\centering
\begin{tabular}{@{}lcrrr@{}}
\toprule
\multicolumn{1}{c}{\textbf{Method}} & \textbf{Layer} & \multicolumn{1}{c}{\textbf{Recall @ 0.7}} & \multicolumn{1}{c}{\textbf{R2}} \\ \midrule
Local random forest & 2010 & 21.22\% & 0.76 \\
SpaceNet model & 2010 & \textbf{99.71\%} & 0.00 \\
CrowdAI model & 2010 & 90.70\% & 0.76 \\
Urban3D model & 2010 & 21.22\% & 0.05 \\
Local U-Net & 2010 & 86.92\% & \textbf{0.93} \\ \midrule
Local random forest & 2020 & 31.26\% & 0.22 \\
SpaceNet model & 2020 & \textbf{99.57\%} & 0.05 \\
CrowdAI model & 2020 & 89.08\% & 0.67 \\
Urban3D model & 2020 & 25.48\% & 0.01 \\
Local U-Net & 2020 & 83.73\% & \textbf{0.84} \\ \bottomrule
\end{tabular}
\caption{Model performance on each layer of imagery over Amman evaluated using the \textbf{test} and \textbf{test count} splits. The SpaceNet, CrowdAI and Urban3D models are pre-trained models from~\cite{jiwani2021semantic} and are \textit{not} fine-tuned. Best values in each column are shown in bold.}
\label{tab:results-main}
\end{table}

\begin{table}[th]
\centering
\resizebox{\linewidth}{!}{%
\begin{tabular}{@{}ccc@{}}
\toprule
\multicolumn{1}{l}{\textbf{Road merging}} & Road class separate & Road class merged \\ \midrule
2010 & 0.710 & 0.728 \\
2020 & 0.674 & 0.687 \\ \midrule
\multicolumn{1}{l}{\textbf{Building buffering}} & Buildings not buffered & Buildings buffered \\ \midrule
2010 & 0.720 & 0.718 \\
2020 & 0.649 & 0.710 \\ \midrule
\multicolumn{1}{l}{\textbf{Weight initialization}} & ImageNet & Random \\ \midrule
2010 & 0.670 & 0.768 \\
2020 & 0.664 & 0.697 \\ \bottomrule
\end{tabular}%
}
\caption{Ablation results showing the average impact of different modeling choices. Values are building class F1 score measured.}
\label{tab:results-ablation}
\end{table}

\section{Case study in Amman}
Jordan has received over 670,000 refugees since the beginning of the Syrian conflict in 2011 \cite{UNHCR} . Although overall violence has declined in the region, poor economic performance and limited basic services in hosting cities exacerbate residents' living conditions and livelihoods. Given that the vast majority of the refugees in Jordan, 83 percent, live in cities instead of camps \cite{UNHCR}, the influx of people has caused severe challenges in maintaining and providing basic infrastructure and public services. However, due to limited planning resources and the excessive number of incoming refugees, our understanding of how Amman has changed within this period is limited. Thus, we use the methods described above to create building footprint layers for the 2010 and 2020 imagery, create density maps, and use these maps to identify growth at an urban scale.

\begin{figure}
    \centering
    \includegraphics[width=1\linewidth]{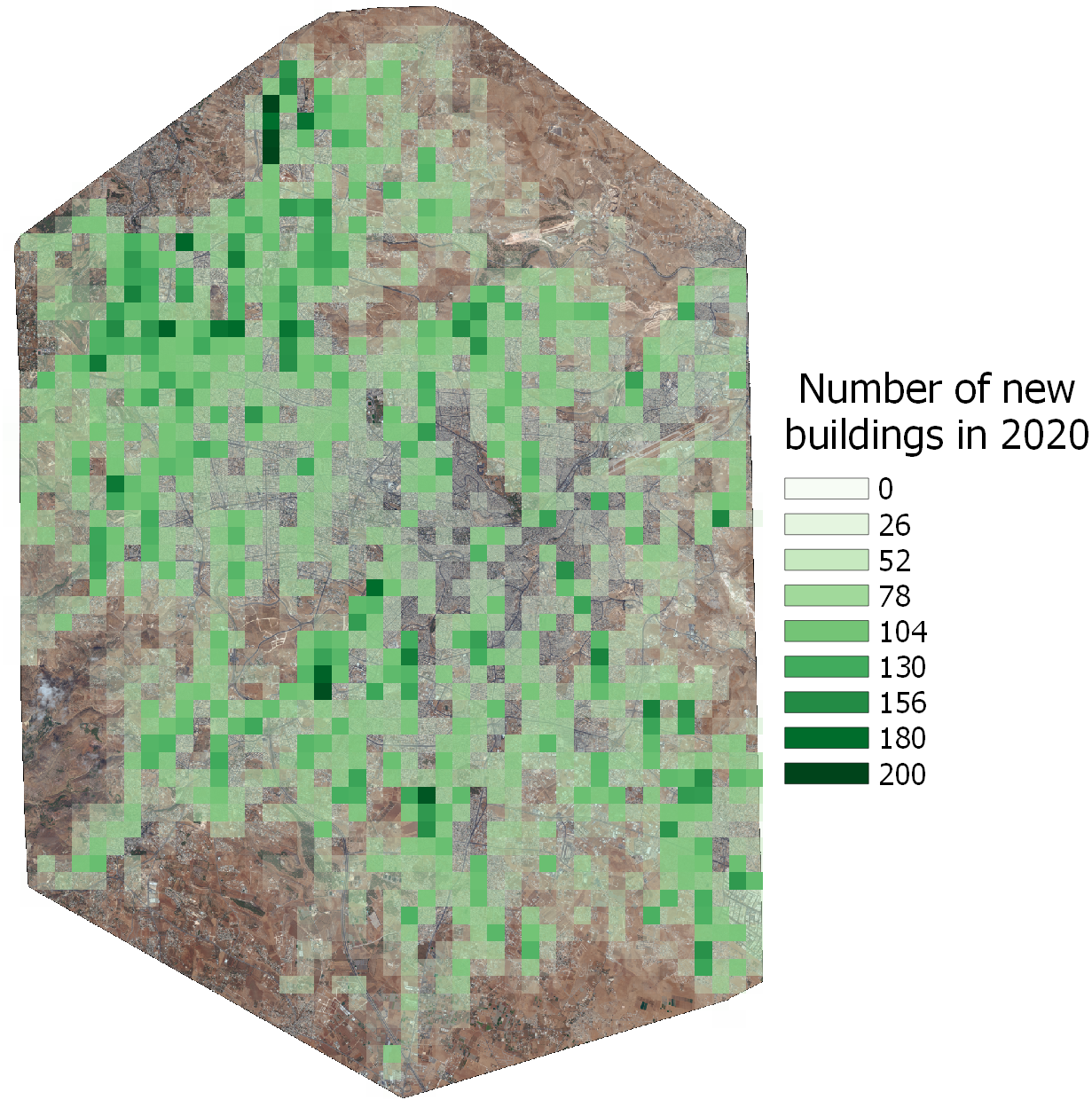}
    \caption{Map of the \textit{increase} in number of buildings between 2010 and 2020 over a 0.25 square kilometer grid in Amman, Jordan.}
    \label{fig:difference-map}
\end{figure}

From Table \ref{tab:results-main} we know that the \textit{count} of buildings predicted by our model are highly correlated with the true count of buildings over $200 \times 200$ meter windows. However, our model could systematically over or under predict buildings and still have highly correlated counts. To account for this and derive counts that accurately reflect the number of buildings on the ground, we can fit a linear regression model using the predicted and actual counts, then adjust our model's predicted counts over the same sized windows using the slope and intercept of the model. After this procedure our models \textit{adjusted} predicted counts have a root-mean-squared error of 2.81 in 2010 and 5.13 in 2020. This allows us to directly compare the counts between the two years of imagery to see \textit{where} in Amman new buildings were created between 2010 and 2020.

Over the entire study area we find $\sim 138,500$ buildings in 2010 and $\sim 150,500$ buildings in 2020, for an estimated growth of 12,000 buildings. Figure \ref{fig:difference-map} shows the spatial distribution of growth, i.e. the difference in number of buildings between 2020 and 2010 over 0.25 square kilometer grid cells. We observe that the north-west portion of the city has experienced the largest amount of growth.

The change in number of buildings over time sheds light on the spatial configuration of the city. In a post-conflict setting, official statistical surveys (i.e., census) often cannot be carried out and/or do not fully address emerging population growth due to limited institutional resources and long-time span of survey instruments. However municipal and central governments still need to allocate fiscal/institutional resources to align with the broader goals of city development (e.g., solid waste management and water and sewer services). Building change maps enable us to understand when/where new buildings were constructed and thus estimate the associated needs of infrastructure and basic service provision. The findings in this case study will be used to carry out strategic policy dialogue with the local and central government.

\paragraph{Low-quality building classification}
We can further investigate the predicted building footprints to understand quality of buildings in the city of Amman. In the context of post-conflict and disaster setting, incoming refugees and migrants are often found living in poor quality housing that needs immediate assistance. Here, we define the quality of buildings based on their morphological features (e.g., size, density, shape, slope, etc.). From our partner's local experience in Amman we know low quality buildings are unevenly distributed within a plot, small, irregularly shaped, often built on steep terrain, and are closely packed with surrounding buildings (i.e., narrow gap between buildings).

\begin{figure}[t!]
\centering
\includegraphics[width=0.9\linewidth]{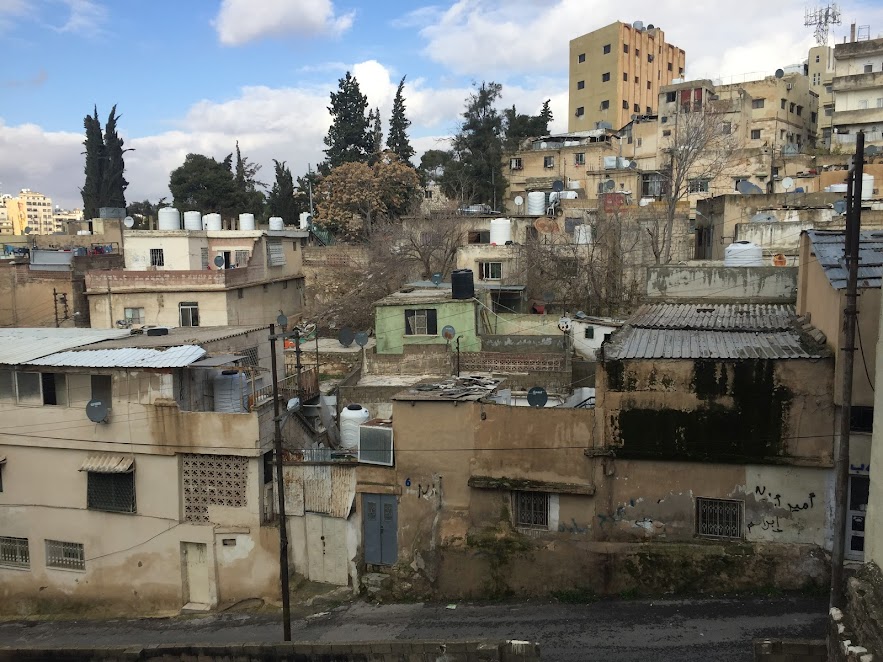}
\caption{Low-quality housing neighborhood in Amman (photo taken on Feb 6, 2022)}
\label{fig:amman}
\end{figure}

First, we extract building level features from the predicted footprint layers, following methodology similar to~\cite{ashilah2021urban,durst2021building}. Specifically, for each building we determine: the area of the footprint, the ratio of the area of the minimum bounding rectangle to the area of the footprint, the number of other buildings in a 200-meter radius, the distance to the nearest building, the maximum slope of the ground the building is on, and the number of corners in the building.

Second, we generate labels indicating whether a building is either ``Regular'' or ``Low-quality'' in two ways: 1) through visual inspections of the high-resolution imagery with a series of consultations with local experts, and 2) through a field visit to take geo-coded ``ground truth'' photos. During the field visit, which was held in Feb 3-8, 2022, we took 308 photos near Al Nathif, Al Akhdar, and Al Zohour. As shown in Figure \ref{fig:amman}, these unplanned areas suffer from overcrowding and limited basic service and infrastructure provisions. Overall, we collected a dataset that positively identifies 399 ``Regular'' buildings and 105 ``Low-quality'' buildings using the 2020 layer of imagery.

\begin{figure}[t!]
\centering
\includegraphics[width=1\linewidth]{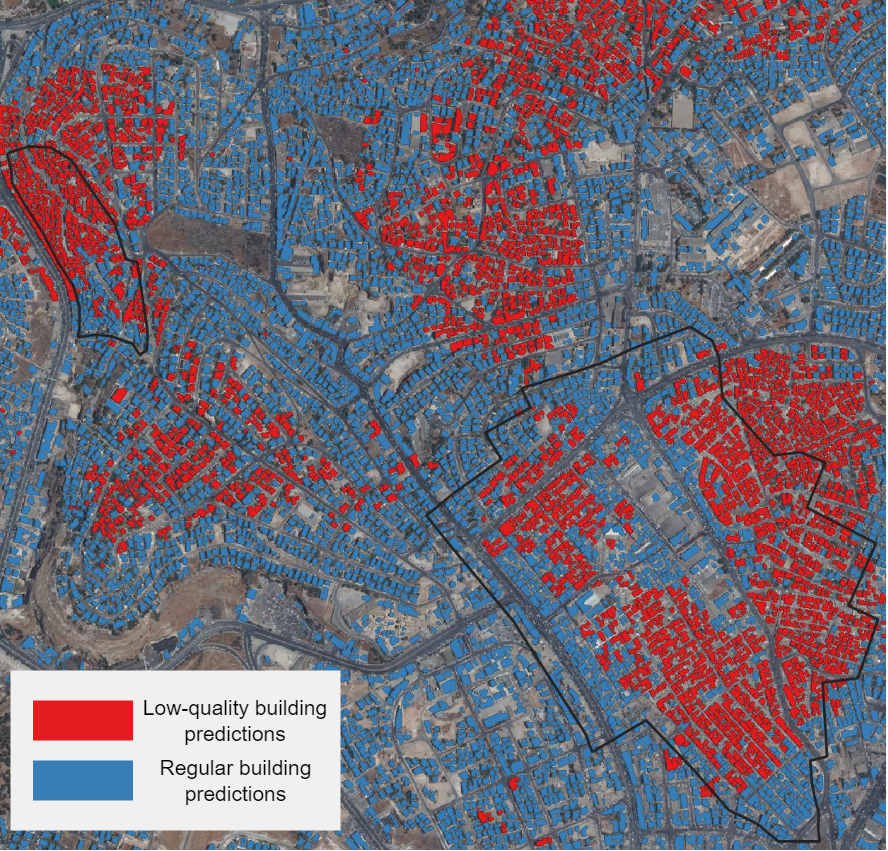}
\caption{Map of predicted building footprints along with their predicted regular or low-quality classification in a section of Amman in 2010. Black outlines show known refugee camp locations. The model predictions align with on-the-ground knowledge.}
\label{fig:formal-informal-map}
\end{figure}

Using the above features and labels, we train a gradient boosting classifier 50 times on random 25\%/75\% data splits and report the average and standard deviation F1 score per class. Here, we observe F1 scores of $0.97 \pm 0.01$ and $0.89 \pm 0.03$ for the regular and low-quality classes respectively. We find that the number of neighbors in 200m of a building is the most important feature in the model, with an importance of 87.23\%, which aligns with how the local experts evaluate the imagery. Importantly, the trained gradient boosting classifier is now able to be run on any layer of predicted footprints (despite only having labels from recent years) after the same feature extraction step. Figure \ref{fig:formal-informal-map}, for example, shows the predictions of the model over part of predicted building footprint layer for 2010. This allows us to further break down the changes in Amman by predicted building quality.

\section{Conclusion}
Building footprint segmentation from high-resolution imagery is an important task in many urban planning and disaster response applications. However, creating models that generalize well for on-demand building footprint segmentation is difficult due to differences in input imagery. We propose a workflow to bypass problems in generalization that simply involves creating labels for a new scene, training a building segmentation model from scratch, then running the model over the scene. We argue that this workflow is easily reproducible in \textit{any} location with several hours of labeling efforts. As an example, we successfully apply this method in a case study mapping buildings over time in Amman, Jordan to quantify urban change.

\section*{Acknowledgements}
This work was made possible by a grant from Microsoft's AI for Humanitarian Action program.

{\small
\bibliographystyle{ieee_fullname}
\bibliography{citations}
}

\end{document}